\documentclass[conf]{new-aiaa}
\usepackage[utf8]{inputenc}

\usepackage{graphicx}
\usepackage{amsmath}
\usepackage[version=4]{mhchem}
\usepackage{siunitx}
\usepackage{longtable,tabularx}
\usepackage[utf8]{inputenc}
\usepackage{graphicx}
\usepackage{amsfonts,amsmath}
\usepackage[version=4]{mhchem}
\usepackage{siunitx}
\usepackage{longtable,tabularx}
\usepackage{adjustbox}
\usepackage{dsfont}
\usepackage{subcaption}
\usepackage{comment}
\usepackage{svg}
\usepackage{gensymb}
\usepackage{multirow}
\usepackage{cleveref}
\graphicspath{{figs/}}
\usepackage{array}
\usepackage{bm}
\usepackage{float}
\usepackage{booktabs}

\newcolumntype{L}[1]{>{\raggedright\let\newline\\\arraybackslash\hspace{0pt}}m{#1}}
\newcolumntype{C}[1]{>{\centering\let\newline\\\arraybackslash\hspace{0pt}}m{#1}}
\newcolumntype{R}[1]{>{\raggedleft\let\newline\\\arraybackslash\hspace{0pt}}m{#1}}

\usepackage{soul}
\definecolor{rev1}{HTML}{FF999A} 
\definecolor{rev2}{HTML}{F3F298} 
\definecolor{rev3}{HTML}{B2E0AE} 
\definecolor{other}{HTML}{C8C7FF} 
\sethlcolor{rev1}

\soulregister\cite7
\soulregister\citep7
\soulregister\citet7
\soulregister\ref7
\soulregister\eqref7

\title{Transformer-Guided Deep Reinforcement Learning for Optimal Takeoff Trajectory Design of an eVTOL Drone}

\author[]{Nathan M. Roberts II~\footnote{Doctoral Student, Department of Mechanical and Aerospace Engineering, AIAA Student Member}}
\author[]{Xiaosong Du~\footnote{Assistant Professor, Department of Mechanical and Aerospace Engineering, AIAA member.}}
\affil[]{Missouri University of Science and Technology, Rolla, MO 65409}

\begin{document}

\maketitle

\begin{abstract}
The rapid advancement of electric vertical takeoff and landing (eVTOL) aircraft offers a promising opportunity to alleviate urban traffic congestion but is still limited by excessive power demands, especially during the takeoff phase. 
Thus, developing optimal takeoff trajectories for minimum energy consumption becomes essential for broader eVTOL aircraft applications. 
Conventional optimal control methods (such as dynamic programming and linear quadratic regulator) provide highly efficient and well-established solutions but are prohibited by problem dimensionality and complexity. 
Deep reinforcement learning (DRL) emerges as a special type of artificial intelligence tackling complex, nonlinear systems; however, the training difficulty is a key bottleneck that hinders DRL applications. 
To address these challenges, we propose the transformer-guided DRL to alleviate the training difficulty by exploring a realistic state space at each time step using a transformer.
The proposed transformer-guided DRL was demonstrated on an optimal takeoff trajectory design of an eVTOL drone for minimal energy consumption while meeting takeoff conditions (\emph{i.e.}, minimum vertical displacement and minimum horizontal velocity) by varying control variables (\emph{i.e.}, power and wing angle to the vertical).
Results presented that the transformer-guided DRL agent learned to take off with $4.57\times10^6$ time steps, representing $25\%$ of the $19.79\times10^6$ time steps needed by a vanilla DRL agent.
In addition, the transformer-guided DRL achieved $97.2\%$ accuracy on the optimal energy consumption compared against the simulation-based optimal reference, while the vanilla DRL achieved $96.1\%$ accuracy. 
Therefore, the proposed transformer-guided DRL outperformed vanilla DRL in terms of both training efficiency and optimal design verification. 
\end{abstract}

\section{Introduction}
\lettrine{E}{lectric} vertical takeoff and landing (eVTOL) aircraft have emerged as an enabling technology for urban air mobility (UAM)~\cite{tripaldi2025emerging, UberElevate, WiskCONOPS, FAACONOPS, Wild2024, Zhou2024}.
UAM systems seek to alleviate urban congestion and improve transportation efficiency around and between metropolitan areas.
eVTOL aircraft are particularly well-suited for densely populated regions due to their precise control, low acoustic footprint, and environmental friendliness~\cite{Andre2019}.
In response to fast-paced advancements in UAM and eVTOL technologies, the Federal Aviation Administration has been working to develop regulations and certification frameworks for drones and small aircraft—including eVTOL vehicles—operating in urban environments~\cite{FAA_AC, FAADrone}, as recognition of and support for this new transportation concept.

Despite the focus and promising advancements, eVTOL technology also presents new challenges, among which energy consumption, especially during the takeoff phase, has become a key bottleneck.
Optimal takeoff trajectory designs for eVTOL aircraft have been investigated to address the energy concerns~\cite{Wang2023, Wei2024, Yeh2024, Sisk2024}.
\citet{Chauhan2020} performed multidisciplinary design optimization on the takeoff trajectory of an eVTOL aircraft to minimize energy use while maintaining passenger comfort with a maximum acceleration constraint.
They also investigated the effects of constraints (\emph{e.g.}, maximum angle of attack) on the optimal takeoff trajectories. 
NASA developed the \texttt{Dymos}~\cite{Falck2021} framework for the control of dynamic, multidisciplinary systems and included the reference~\cite{Chauhan2020} as a verification case. 

Recently, artificial intelligence methods have been applied to eVTOL takeoff trajectory optimization to improve energy efficiency~\cite{Du2026}.
\citet{Sisk2024} developed twin-generator generative adversarial networks to generate realistic eVTOL takeoff trajectories.
The developed model facilitated surrogate modeling through the generation of realistic trajectories and dimensionality reduction, enabling efficient takeoff trajectory optimization. 
\citet{Roberts2025} developed a deep reinforcement learning (DRL) framework with an effective reward function formulation to learn the optimal takeoff policy.
The trained policy achieved greater than 96\% accuracy for optimal energy consumption, but we noticed high sample inefficiency.

Transformers are a family of neural network architectures developed for natural language processing tasks, where the attention-based mechanism performs well in translation and sequence understanding~\cite{Attention}.
Unlike recurrent architectures (\emph{e.g.}, recurrent neural networks and long short-term memory networks), transformers process sequential data in parallel, efficiently learning long-range relationships using self-attention.
The ability of a transformer to handle sequential data has led to applications beyond language processing, including time-series forecasting~\cite{wen2022}.
In this work, we propose to use a transformer to represent the temporal changes in eVTOL control profiles.
We use the trained transformer to guide a DRL agent operating in a complex, nonlinear environment.
By integrating transformer-based sequential modeling with DRL, the proposed framework is expected to improve policy convergence for DRL and energy efficiency for eVTOL takeoff trajectory optimization.

The remainder of this paper is organized as follows.
In Sec.~\ref{sec:methodology}, we describe the eVTOL simulation models, transformer architecture, DRL framework, and transformer-guided DRL approach.
In Sec.~\ref{sec:results}, we present the results achieved by the transformer-guided DRL compared against simulation-based optimal references and vanilla DRL-based results.
We end this paper with conclusions and future work in Sec.~\ref{sec:conclusion}.

\section{Methodology}
\label{sec:methodology}

\subsection{eVTOL Models}
\label{sec:dynamics}
The aircraft considered in this paper is a tandem, tilt-wing, eVTOL aircraft, which is modeled after the Airbus $A^3$ Vahana~\cite{AirbusA3}.
Each wing has a rectangular planform with a span of \unit{\qty{6}\meter} and a chord length of \unit{\qty{1.5}\meter}, using the NACA~0012 airfoil as the cross-sectional profile.
Each wing has four three-bladed propellers with a radius of \unit{\qty{0.75}\meter} and a blade chord of \unit{\qty{0.1}\meter}.
{The front and rear wings are assumed to be geometrically identical and rotate in sync, balancing the pitch moments, removing the need to explicitly model trim and stability~\cite{Chauhan2020}.
This symmetry reduces the degree of freedom to vertical and horizontal displacements only, which are measured with respect to a fixed inertial reference frame.}
The propeller thrust line is assumed to be in line with the wing chord.
The body of the aircraft is assumed to have a drag area of \unit{\qty{0.35}\meter\squared}, and the related drag coefficient does not vary with the angle of attack.
The \texttt{Dymos} framework~\cite{Falck2021} developed by NASA is used to generate optimal design references. 

\subsubsection{Aerodynamics}
\label{sec:Aerodynamics}
Pre-stall aerodynamic coefficients are provided using finite-wing corrections with lifting-line theory.
While transitioning from vertical to horizontal flight, a wing may experience stall conditions where linear aerodynamic assumptions are no longer valid.
Lift and drag beyond the linear region must be available to accurately model this transition.
{\citet{Tangler1991} used experimental data} to create an empirical model, estimating the post-stall lift coefficient~(Eqn.~\ref{eq:1}) and drag coefficient~(Eqn.~\ref{eq:2}) for a finite-length rectangular wing.
\begin{equation}
    \label{eq:1}
    C_L = A_1 \sin{(2\alpha)} + A_2 \frac{\cos{(\alpha)}^2}{\sin{(\alpha)}}\text{,}
\end{equation}
where
\begin{equation}
    A_1 = \frac{C_1}{2}\text{,}
\end{equation}
\begin{equation}
    A_2 = \left(C_{L_s} - C_1 \sin{(\alpha_s)}\cos{(\alpha_s)}\right)\frac{\sin{(\alpha_s)}}{\cos{(\alpha_s)}^2}\text{,}
\end{equation}
\begin{equation}
    C_1 = 1.1 + 0.018AR\text{,}
\end{equation}
$\alpha$ is the wing angle of attack, $\alpha_s$ is the angle of attack at the stall, $C_{L_s}$ is the lift coefficient at stall, and $AR$ is the wing aspect ratio. 
For post-stall angles of attack above $27.5$ degrees, the drag coefficient is computed as follows.
\begin{equation}
    \label{eq:2}
    C_D = B_1 \sin{(\alpha)} + B_2 \cos{(\alpha)}\text{,}
\end{equation}
where
\begin{equation}
    B_1 = C_{D_{max}}\text{,}
\end{equation}
\begin{equation}
    B_2 = \frac{C_{D_s} - C_{D_{max}} \sin{(\alpha_s)}}{\cos{(\alpha_s)}}\text{,}
\end{equation}
\begin{equation}
    C_{D_{max}} = \frac{1.0 + 0.065AR}{0.9 + t/c}\text{,}
\end{equation}
$C_{D_s}$ represents the drag coefficient at stall, and $t/c$ represents the thickness-to-chord ratio of the airfoil.
For angles of attack below $27.5$ degrees, the aerodynamic quantities experimentally gathered by~\citet{Tangler1991} are used~({Table}~\ref{tab:1}).
\citet{Tangler1991} showed that the airfoil geometry had a minimal effect on the post-stall characteristics, so the geometry is not included in~Eqns.~\ref{eq:1} and~\ref{eq:2}.
\citet{Tangler1991} also showed that the Reynolds number has little effect on the post-stall characteristics within the range of $0.25 \times 10^6$ to $1 \times 10^6$.
This range is reasonable for the takeoff phase of the aircraft under consideration.
Additionally, while the wings are assumed to be independent when calculating the lift and parasite drag, the induced drag is affected by aerodynamic interaction between the wings due to the tandem biplane configuration.
Due to the wingtip motors, the span efficiency is assumed to be 0.95.
The induced drag is determined as 
\begin{equation}
    \label{eq:3}
    D_{induced} = \frac{1}{\pi q}\left(2\frac{L^2_{wing}}{b^2}+2\sigma\frac{L^2_{wing}}{b^2}\right),
\end{equation}
where $q$ is the dynamic pressure and $b$ is the span.
For a gap-to-span ratio of 0.25, $\sigma$ is approximately 0.4.
This simplifies to the following, where it can be seen that the tandem wings reduce the span efficiency of both wings to 0.68, which is used to calculate the lift and drag for each wing independently.
\begin{equation}
    \label{eq:4}
    D_{induced} = 2\left(\frac{L^2_{wing}}{0.68 \cdot \pi \cdot q \cdot b^2}\right).
\end{equation}

Since the lift and drag curves are not smooth where the models change, Kreisselmeier--Steinhauser (KS) functions are used to make them smooth to support gradient-based optimization.

\begin{table}[h]
    
    \caption{Drag coefficient data points between 16 and 27.5 degrees from \citet{Tangler1991}.}
    \label{tab:1}
    \newcolumntype{C}{>{\centering\arraybackslash}X}
\begin{tabularx}{\textwidth}{C C C C C}
\toprule
        \boldmath{$\alpha$} & \textbf{16}\unit{\degree} & \textbf{20}\unit{\degree} & \textbf{25}\unit{\degree} & \textbf{27.5}\unit{\degree} \\
    \midrule
        $C_D$ & 0.100 & 0.175 & 0.275 & 0.363 \\
\bottomrule
    \end{tabularx}
\end{table}

\subsubsection{Propulsion}
The thrust, as a function of power provided, is calculated using momentum theory as follows.
\begin{equation}
    \label{eq:5}
    P_{disk} = TV_{\infty \perp} + \kappa T \left(-\frac{V_{\infty \perp}}{2} + \sqrt{\frac{V^2_{\infty \perp}}{4} + \frac{T}{2\rho A_{disk}}}\right),
\end{equation}
where $P_{disk}$ represents the power supplied to the propeller disk, $T$ represents the thrust, $V_{\infty \perp}$ represents the normal component of the freestream velocity, $\rho$ represents the air density, $A_{disk}$ represents the area of the propeller disk, and $\kappa = 1.2$ represents the correction factor accounting for nonuniform inflow, tip effects, and simplifications from momentum theory.
The thrust must be solved numerically, since the power provided is a design variable.
The Newton-Raphson method is applied with an initial guess of 1.2 times the weight of the aircraft.

The profile power can be estimated as
\begin{equation}
    \label{eq:6}
    C_{P_p} = \frac{\sigma C_{d_{0_p}}}{8} (1 + 4.6 \mu^2),
\end{equation}
where
\begin{equation}
    \mu = \frac{V_{\infty \parallel}}{\Omega R},
\end{equation}
\begin{equation}
    C_{P_p} = \frac{P_p}{\rho A_{disk}R^3\Omega^3},
\end{equation}
where $\sigma$ represents the solidity, $C_{d_{0_p}}$ represents the constant profile drag coefficient, $V_{\infty \parallel}$ represents the parallel component of the freestream velocity, and $C_{P_p}$ represents the coefficient of profile power.
It is assumed that the propellers are variable-pitch and maintain a constant $\Omega =$ \unit[per-mode=symbol]{\qty{181}\radian\per\second}.
The solidity is calculated to be 0.13, and $C_{d_{0_p}}$ is assumed to be 0.012.
The power supplied to the propeller disk can be calculated as follows: $P_{disk} = \eta P_{electrical} - P_p$,
with the supplied electric power as input.
A loss factor $\eta$ is used to account for mechanical and electrical losses.

The propellers will produce normal forces when the freestream flow is not aligned with the propeller axes.
This force can be approximated using the empirical formula from~\citet{DeYoung1965} as follows:
\begin{equation}
    \label{eq:8}
    N = \frac{4.25 \sigma_e \sin{(\beta + 8\degree)}fq_{\perp}A_{disk}}{1+2\sigma_e} \tan{(\alpha_{in})},
\end{equation}
where
\begin{equation}
    \sigma_e = \frac{2Bc_b}{3\pi R},
\end{equation}
\begin{equation}
    f = 1 + \frac{\sqrt{1+T_c} - 1}{2} + \frac{T_c}{4(2+T_c)},
\end{equation}
\begin{equation}
    T_c = \frac{T}{q_{\perp} A_{disk}},
\end{equation}
$\beta$ represents the blade pitch angle at three-quarters of the radius, $q_{\perp}$ represents the dynamic pressure based on the normal component of the velocity, $A_{disk}$ represents the propeller disk area, $\alpha_{in}$ represents the incidence angle, $\sigma_e$ represents the effective solidity, $B$ represents the number of blades per propeller, $c_b$ represents the average blade chord, $R$ represents the propeller radius, $f$ represents the thrust factor, $T_c$ represents the thrust coefficient, and $T$ represents the thrust.
$\beta$ is assumed to change linearly from 0 to 35\unit{\degree} as the flight speed increases from 0 to \unit[per-mode=symbol]{\qty{67}\meter\per\second}.

\subsubsection{Propeller-Wing Interaction}
Momentum theory is used to model the interaction between the induced flow from the propellers to the wings.
Momentum theory finds a theoretical upper limit ($2v_i$) to the amount the flow speed can increase when it passes through a propeller disk. The induced velocity, $v_i$, can be computed as follows:
\begin{equation}
    \label{eq:9}
    v_i = \left(-\frac{V_{\infty \perp}}{2} + 2\sqrt{\frac{V^2_{\infty \perp}}{4} + \frac{T}{2 \rho A_{disk}}}\right)\text{.}
\end{equation}

\citet{Chauhan2020} calculated the propeller-wing interaction as $V_{\infty \perp} + k_w v_i$ and tested a range of $k_w$ between 0 and 200\%.
They showed that the propeller-wing interactions had a negligible effect on the power required for takeoff.
Therefore, we set $k_w = 1$ in this work. 

\subsubsection{Dynamics}
The $x$ and $y$ components of the velocity at the next step are given by 
\begingroup\makeatletter\def\f@size{9}\check@mathfonts
\def\maketag@@@#1{\hbox{\m@th\fontsize{10}{10}\selectfont\normalfont#1}}
\begin{equation}
    \label{eq:10}
    v_{x_{i+1}} = v_{x_i} + \frac{T_i\sin{(\theta_i)} - D_{fuse_i}\sin{(\theta_i+\alpha_{\infty_i})} - D_{wings_i}\sin{(\theta_i + \alpha_{EFS_i})} - L_{wings_i}\cos{(\theta_i+\alpha_{EFS_i})} - N_i\cos{(\theta_i)}}{m}\delta t
\end{equation}
\endgroup
and
\begingroup\makeatletter\def\f@size{9}\check@mathfonts
\def\maketag@@@#1{\hbox{\m@th\fontsize{10}{10}\selectfont\normalfont#1}}
\begin{equation}
    \label{eq:11}
    v_{y_{i+1}} = v_{y_i} + \frac{T_i\cos{(\theta_i)} - D_{fuse_i}\cos{(\theta_i+\alpha_{\infty_i})} - D_{wings_i}\cos{(\theta_i + \alpha_{EFS_i})} - L_{wings_i}\sin{(\theta_i+\alpha_{EFS_i})} + N_i\sin{(\theta_i)} -mg}{m}\delta t,
\end{equation}
\endgroup
where $i$ is the index of the time step, $\delta t$ is the length of each time step, $\theta$ represents the wing angle relative to vertical, $\alpha_{\infty}$ represents the freestream angle of attack, $\alpha_{EFS}$ represents the effective freestream angle of attack due to the propellers, $m$ represents the mass of the aircraft, $T$ represents the total thrust, $D_{fuse}$ represents the drag of the fuselage, $D_{wings}$ represents the total drag of both wings, $L_{wings}$ represents the total lift of both wings, and $N$ represents the total normal force from the propellers.
The forward Euler method is used to propagate the aircraft states.

\subsection{Transformer-Guided Deep Reinforcement Learning}
\label{sec:TDRL}

\subsubsection{Transformer}
\label{sec:Trans}
The transformer relies on a self-attention mechanism that enables each element in a sequence to attend to the other elements simultaneously, without relying on recurrence~\cite{Attention}.
The parallel computation allows transformers to learn sequential or temporal relationships more efficiently than recurrent neural networks or long short-term memory networks, which process sequences step by step.
In this work, we extend the transformer to model the temporal evolution of control variables.
The transformer is trained on optimal trajectories generated using \texttt{Dymos}, representing the underlying temporal patterns of the optimal control profiles.

The self-attention mechanism of transformers works by focusing on relevant portions of a sequence while ignoring less relevant portions.
Positional encoding integrates relative positions into a sequence.
The relevance of one element in a sequence to other elements is determined using query, key, and value vectors.
For a sequence of length $n$ consisting of $d$-dimensional vectors $\bm{X}\in\mathbb{R}^{n\times d}$, the positional encoding is typically done using sine and cosine functions:
\begin{equation}
    \label{eq:pe}
    \bm{X}_{enc}=\bm{X}+\bm{PE}\text{,}
\end{equation}
where
\begin{equation*}
    \bm{PE}_{pos,2i}=\sin\left(\frac{pos}{10000^{\frac{2i}{d}}}\right)\text{,}
\end{equation*}
\begin{equation*}
    \bm{PE}_{pos,2i+1}=\cos\left(\frac{pos}{10000^\frac{2i}{d}}\right)\text{,}
\end{equation*}
where $pos$ is the position in the sequence, and $i$ is the dimension.
The query vector represents how each element evaluates the relevance of all elements—including itself—to itself.
The query vector for an element is calculated by multiplying the encoded sequence by a trainable weight matrix, $\bm{W}_q\in\mathbb{R}^{d\times d_k}$, where $d_k$ is a hyperparameter:
\begin{equation}
    \label{eq:query}
    \bm{q}_i=\bm{x}_{\text{enc},i}\bm{W}_q\in\mathbb{R}^{1\times d_k}\text{.}
\end{equation}
Each element encodes its relevance using a key vector, which is similarly calculated using a trainable weight matrix:
\begin{equation}
    \label{eq:key}
    \bm{k}_i=\bm{x}_{\text{enc},i}\bm{W}_k\in\mathbb{R}^{1\times d_k}\text{.}
\end{equation}
The value vector is created in the same manner and represents the information a particular element can offer:
\begin{equation}
    \label{eq:value}
    \bm{v}_i=\bm{x}_{\text{enc},i}\bm{W}_v\in\mathbb{R}^{1\times d_k}\text{.}
\end{equation}

An attention vector is calculated for an element by taking the dot product of the query vector with every key vector, measuring the compatibility of the query and key vectors.
\begin{equation}
    \label{eq:attention}
    \bm{a}_i=\text{softmax}\left(\frac{\left[\bm{q}_i^\top\bm{k}_1, \bm{q}_i^\top\bm{k}_2,\ldots,\bm{q}_i^\top\bm{k}_n\right]}{\sqrt{d_k}}\right)^\top\in\mathbb{R}^{1\times n}\text{.}
\end{equation}
Elements of an attention vector will be close to zero if the corresponding key vectors are dissimilar to the query vector---meaning the corresponding vector of sequence $\bm{X}_{enc}$ is irrelevant to $\bm{x}_{enc,i}$---and close to one if they are similar.
The output of the attention mechanism is a weighted sum of the value vectors, using an element's attention vector as the weights,
\begin{equation}
    \label{eq:outputs}
    \bm{z}_i=\sum_{j=1}^{n}{\bm{a}_{ij}\bm{v}_j}\in\mathbb{R}^{1\times d_k}\text{.}
\end{equation}
In practice, the attention mechanism is simultaneously run for many queries that are stacked into $\bm{Q}$, with the keys and values becoming $\bm{K}$ and $\bm{V}$, respectively.
The matrix of outputs is then computed as
\begin{equation}
    \label{eq:mechanism}
    \text{Attention}\left(\bm{Q},\bm{K},\bm{V}\right)=\text{softmax}\left(\frac{\bm{Q}\bm{K}^\top}{\sqrt{d_k}}\right)\bm{V}\in\mathbb{R}^{n\times d_k}\text{.}
\end{equation}

In applications where the transformer will be generating one element at a time, as is the case in this work, it is important to prevent later elements in a trajectory from being attended to.
This is done while calculating the attention matrix by masking out the inputs to the softmax containing unwanted connections, as shown below:
\begin{equation}
    \label{eq:mask}
    \bm{A}=\text{softmax}\left(\frac{\bm{Q}\bm{K}^\top}{\sqrt{d_k}}+\bm{M}\right)\text{,}
\end{equation}
where the mask is a strictly upper triangular matrix of size $n$,
\begin{equation*}
    \bm{M}=\begin{bmatrix}
    0 & -\infty & -\infty & \cdots & -\infty \\
    0 & 0 & -\infty & \cdots & -\infty \\
    0 & 0 & 0 & \cdots & -\infty \\
    \vdots & \vdots & \vdots & \ddots & \vdots \\
    0 & 0 & 0 & \cdots & 0
    \end{bmatrix}\in\mathbb{R}^{n\times n}\text{.}
\end{equation*}

To capture multiple representations or underlying relationships, it is common to perform multiple separate attention operations on the same set of queries, keys, and values.
Multi-head attention projects the queries, keys, and values with additional learnable matrices, then attends to each of the projections, and linearly combines the outputs
\begin{equation}
    \label{eq:multihead}
    \text{MultiHead}\left(\bm{Q},\bm{K},\bm{V}\right)=\text{Concat}\left(\text{head}_1,\text{head}_2,\ldots,\text{head}_h\right)\bm{W}_O\in\mathbb{R}^{n\times d_v}\text{,}
\end{equation}
where
\begin{equation*}
    \text{head}_i=\text{Attention}\left(\bm{Q}\bm{W}_{Q,i},\bm{K}\bm{W}_{K,i},\bm{V}\bm{W}_{V,i}\right)\text{,}
\end{equation*}
$\bm{W}_{Q,i},\bm{W}_{K,i}\in\mathbb{R}^{d\times d_k}$, $\bm{W}_{V,i}\in\mathbb{R}^{d\times d_v}$, and $\bm{W}_O\in\mathbb{R}^{d_vh\times d}$ are learnable parameters, while the value dimension $d_v$ and number of heads $h$ are hyperparameters.

In this study, a decoder-only transformer is trained on optimal control trajectories to learn the temporal relations within the optimal control profiles.
Each training step is conducted as follows:
\begin{enumerate}
    \item A batch of trajectories is pulled from the training set.
    \item Because the trajectories have low-dimensional elements (\emph{i.e.}, two control actions at each time step in this work), they are projected into a higher-dimensional space using a single-layer feed-forward neural network (FNN) such that $d=d_k=d_v=64$.
    \item The projected trajectories are positionally embedded following the procedure described in Eqn.~\ref{eq:pe}.
    \item The embedded trajectories are processed through two masked multi-head self-attention layers.
    \item The attention outputs are passed through another single-layer FNN to create distribution parameters: mean $\mu$ and variance $\sigma^2$.
    \item The negative log-likelihood (Eqn.~\ref{eq:nll}) of producing the next element under the distribution described by the parameters serves as a loss function,
\begin{equation}
    \label{eq:nll}
    \mathcal{L}_\text{NLL}=\frac{1}{2N}\sum_{i=1}^N\sum_{j=1}^D\left[\ln(2\pi\sigma_{i,j}^2)+\frac{(y_{i,j}-\mu_{i,j})^2}{\sigma_{i,j}^2}\right]\text{,}
\end{equation}
where $N$ is the total number of samples in the batch, and $D$ is the dimensionality of the sequence elements.
    \item The Adam optimizer takes one step.
\end{enumerate}

\subsubsection{Deep Reinforcement Learning}
\label{sec:DRL}

Reinforcement learning (RL) is a machine learning framework in which an agent interacts with an environment, learning to execute optimal actions for a given state, with the goal of maximizing cumulative rewards over an episode~\cite{Kaelbling1996}.
RL is modeled as a Markov decision process, which consists of a set of possible states $S$, called the state space, a set of possible actions $A$, called the action space, the probability $P_{a}\left(s,s'\right)$  of transitioning from state $s$ to state $s'$ given action $a$, and the reward $R_{a}\left(s,s'\right)$ for taking action $a$ to transition from state $s$ to $s'$.

An RL agent acting at time $t$ takes the current state $S_t$ as input and samples action $A_t$ from the learned distribution $\pi\left(S_t\right)$.
The state then transitions to $S_{t+1}$ with probability $P\left(S_{t+1}|A_t,S_t\right)$, and a reward $R_{t+1}$ is received. 
The policy $\pi$ is a mapping of the state space to the action space as follows,
\begin{equation}
    \label{eq:policy}
    \begin{split}
        &\pi\left(a|s\right)=\text{P}\left(A_t=a|S_t=s\right)\text{.}
    \end{split}
\end{equation}
The agent aims to learn the optimal policy $\pi^*(s)$ that maximizes the cumulative reward, $\sum_{t=1}^{N}R_t$, over $N$ time steps.
RL that uses deep neural networks to learn this mapping is known as DRL.

In this work, we use the soft actor-critic (SAC) method~\cite{Haarnoja2018}, in which a policy network (termed actor) selects actions while critic networks (termed critic) evaluate how good those actions are by estimating the value, which is the sum of the immediate reward and cumulative future rewards.
The value networks in the critic predict the expected return of being in state $s$ (\emph{i.e.}, state value), while Q-networks predict the expected return of taking action $a$ in state $s$ (\emph{i.e.}, action value).
The critic is trained using the temporal-difference (TD) error, which reflects the discrepancy between predicted and target returns.
A distinguishing feature of SAC compared with conventional actor-critic methods is the inclusion of an entropy regularization term into the loss function for training the policy~\cite{SpinningUp}
\begin{equation}
    \label{eq:sac}
    L=-Q\left(\tilde{a}_\theta\left(s\right),s\right)+\alpha\log\pi_\theta\left(\tilde{a}_\theta\left(s\right)|s\right)\text{,}
\end{equation}
where $Q\left(\tilde{a}_\theta\left(s\right),s\right)$ is the action value predicted by the Q-network, $\alpha$ is an entropy-regularization parameter, and $\tilde{a}_\theta\left(s\right)$ is an action sampled from $\pi_\theta(s)$.
Entropy, which quantifies the randomness of a probability distribution, is used to encourage exploration by promoting stochasticity in the policy.
Importantly, SAC operates off-policy, meaning that the data used to train the networks does not need to be generated by the current policy.
Instead, experiences collected from previous iterations of the policy can still contribute to improving the performance of the actor and critic networks.

The SAC agent in this work is trained for $5\times10^6$ time steps, implemented via the \texttt{Stable~Baselines3} library~\cite{stable-baselines3}.
The agent is evaluated every $1\times10^4$ steps to save the model with the highest cumulative reward.
Both the actor and critic networks are composed of three fully connected layers, each containing $512$ neurons and employing the \texttt{ReLU} activation function.
Training is conducted using a linearly annealed learning rate starting at $4\times10^{-4}$, a batch size of $256$, a soft target update coefficient of $5\times10^{-3}$, a discount factor of $0.98$, and an entropy coefficient of $0.01$.
To mitigate issues such as catastrophic forgetting and policy collapse, the agent utilizes a replay buffer capable of storing up to $5\times10^6$ elements.

\subsubsection{Transformer-Guided Deep Reinforcement Learning}
\label{sec:sTDRL}

A major drawback of DRL is its requirement for large amounts of training data and simulation time.
Environments with strict constraints or early termination criteria are particularly prone to this issue.
eVTOL takeoff trajectory design, due to the dangers inherent in air travel, necessitates such constraints.
To address the training difficulty, we propose the transformer-guided DRL for optimal takeoff trajectory design in this work.

\begin{figure}[H]
    \centering
    \includegraphics[width=0.5\linewidth]{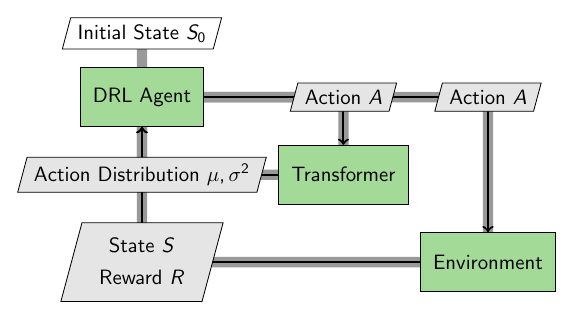}
    \caption{The training loop of a transformer-guided DRL agent.}
    \label{fig:tdrl}
\end{figure}

The transformer-guided DRL (Fig.~\ref{fig:tdrl}) selects an action from a distribution that the trained transformer generates, while vanilla DRL learns to take actions within a pre-defined space, which typically incurs the training difficulty.
Specifically, the transformer produces an action proposal distribution characterized by a mean and variance for each action component, conditioned on the previous action history.
The transformer-generated action proposal distribution constrains the effective action space explored by the DRL agent at each time step.
The agent generates a value within [-1, 1], corresponding to the z-score of the action from the distribution that is to be executed.
This range, $\left[-1, 1\right]$, is chosen because it captures the majority of the distribution while keeping the executed action near the mean.
This method substantially reduces the effective exploration space.

After each time step, the executed action is appended to the action history, which is then used by the transformer to generate the next action proposal distribution.
Because the action proposal distribution differs from the executed action, a traditional encoder-decoder transformer architecture is not feasible.
Appending the action distribution parameters to the state space allows the policy to account for previous actions, interpreted through the lens of the transformer.
To allow the DRL agent to select the first action, the initial state from the environment is appended with a start token outside the transformer's output range.
This enables the agent to determine an initial action before transformer guidance becomes available.


\section{Results and Discussion}
\label{sec:results}

\subsection{Problem Formulation}
The optimal takeoff trajectory design problem is formulated for minimum energy consumption with constraints on takeoff conditions~(Table~\ref{tab:opt_setup}). 
Specifically, we define takeoff as reaching an altitude of at least 305 m with a horizontal velocity of at least 67 m/s.
The aircraft cannot go through solid ground, so the altitude has a lower bound of $0$ meters.
The design variables include 40 Gauss-Lobatto collocation points (\emph{i.e.}, 20 for power control profile and 20 for wing angle control profile) and the total takeoff time. 
The reference results are generated using \texttt{Dymos}~\cite{Falck2021}.
\begin{table*}[htbp]
  \small
  \centering
  \caption{\small Takeoff trajectory optimization problem formulation.}
  \label{tab:opt_setup}
  \begin{tabularx}{\textwidth}{@{\extracolsep{\fill}} llllll}
  \toprule
  &  \textbf{Function or Variable}   &  \textbf{Description}   &  \textbf{Quantity} \\
  \midrule   
  minimize   &  $E$ &  Electrical energy consumed &   \\
  \midrule
  w.r.t. & $\bf{P}$ & Electrical power  &  20  \\
         & $\boldsymbol{\theta}$ & Wing angle to vertical  & 20 \\
         & $T_{\text{takeoff}}$ &  Total takeoff time  &  1  \\
         &  & {\bf Total design variables}  &  {\bf 41 }  \\
  \midrule
  subject to & $y_{\text{final}}\geq{305}$ \unit{\meter} &  Final vertical displacement constraint &  1  \\
             & $V_{x,\text{final}} \geq 67$ \unit[per-mode = symbol]{\meter\per\second} & Final horizontal speed constraint & 1 \\
             & $y\geq{0}$ \unit{\meter} & Vertical displacement constraint & 1 \\
             &    & {\bf Total constraints} &  {\bf 3}  \\
    \bottomrule
    \end{tabularx}
\end{table*}

\subsection{Transformer-Guided Deep Reinforcement Learning}

The transformer is trained on $1,000$ reference optimal trajectories~\cite{Yeh2023} gathered via Latin hypercube sampling for five flight condition parameters~(Table~\ref{tab:des_space}).
Specifically, $750$ samples are used to train the transformer, $150$ are used as validation data, and $100$ are used as testing data.
Each of the trajectories is interpolated into $0.1$ second intervals using cubic splines, following the process \texttt{Dymos} uses for time-series outputs.
The transformer is trained based on the true next element under the generated distribution (Sec.~\ref{sec:TDRL}).
In the verification case, we use $k_w=1$, $\eta=0.9$, and $S_\text{ref} = 1$, with only takeoff conditions as shown in Table~\ref{tab:opt_setup}. 
\begin{table}[H]
\caption{Flight-condition space used for generating optimal training trajectories for a transformer. The parameters with subscripts "max" denote corresponding upper bounds.}
     \centering
     \begin{tabular}{l|c c c c c}
     \toprule
         Quantity & $\alpha_{\text{max}}$ & $a_{\text{max}}$ & $k_{w}$ & $\eta$ & $S_{\text{ref}}$\\
         Description & Angle of attack & Acceleration & Induced velocity factor & Efficiency factor & Planform area scale\\
         Range & $\left[10, 15\right]$\unit{\degree} & $\left[0.2, 0.4\right]g$ & $\left[0.3, 1\right]$ & $\left[0.7, 0.9\right]$ & $\left[0.9, 1\right]$\\
        \bottomrule
     \end{tabular}
     \label{tab:des_space}
\end{table}

The decoder-only transformer used in this work has an input dimension $D$ of $2$, a hidden dimension $d_k=d_v$ of $64$, $4$ attention heads, and $2$ attention layers, and utilizes a dropout of $0.1$ during training.
The model is trained using the \texttt{Adam} optimizer with a learning rate of $10^{-4}$ and batch size of $64$ for $100$ epochs.
The best model is saved according to the validation data.
We use the best model to auto-regressively generate trajectories starting from the first action of each of the testing trajectories to evaluate the transformer's ability to generate a realistic or suboptimal trajectory.
The trained transformer achieved $95.4\%$ accuracy (formulated as follows) on energy consumption compared with the testing trajectories.
\begin{equation}
  \label{eq:accuracy}
  \zeta_\text{rel}=\left(1-\frac{E_\text{generated}-E_\texttt{reference}}{E_\texttt{reference}}\right)\times100\%\text{.}
\end{equation}

In this work, we develop a DRL-compatible environment, containing the simulation models, optimization objective, and takeoff conditions, using the \texttt{Gymnasium} interface in \texttt{python}~\cite{gymnasium}.
At the core of DRL lies the reward function, which leads the behavior of the agent.
At each time step within the developed environment, the agent gets feedback composed of several reward components (Table~\ref{tab:rewards}).
The relative weight afforded to each portion of the reward is governed by the hyperparameter $\rho=2$.
To lead the agent toward takeoff, the reward includes a penalty based on the difference between the current state and the target takeoff state.
For effective guidance, these reward components are formulated as convex shapes with singularities~\cite{Huang2022}, with concavity governed by a concavity coefficient hyperparameter $cc=0.5$.
Energy efficiency is promoted by penalizing energy consumption at each time step, encouraging a minimum-energy trajectory.
All rewards are scaled using the scale hyperparameter $k=0.05$.
The negative reward structures incentivize the agent to terminate the episode as soon as possible, which occurs if the agent violates any physical constraints, exceeds the time limit, or successfully takes off.
To prevent the agent from exploiting early termination through constraint violation, the unweighted maximum future energy penalty, $R_{P,\text{max}}\left(t\right)=-10k\left(t_\text{max}-t\right)$, is applied if a flight constraint is violated.
The early termination conditions include hitting the ground or generating a negative freestream velocity.

\begin{table}[h]
\caption{Reward components.}
    \centering
    \begin{tabular}{l|c c c}
    \toprule
    Quantity & $R_y$ & $R_{v_x}$ & $R_P$ \\
    Value & $\frac{ke^{cc}\left(e^{-cc\left|1-\frac{y}{305}\right|}-1\right)}{2\left(e^{cc}-1\right)\left(\rho+1\right)}$ & $\frac{ke^{cc}\left(e^{-cc\left|1-\frac{V_x}{67}\right|}-1\right)}{2\left(e^{cc}-1\right)\left(\rho+1\right)}$ & $\frac{-\rho kP}{310000\left(\rho+1\right)}$ \\
    \bottomrule
    \end{tabular}
    \label{tab:rewards}
\end{table}

The state space is comprised of altitude, horizontal velocity, vertical velocity, mean power, mean wing angle, power variance, and wing angle variance~(Table~\ref{tab:state}).
These quantities are selected because they directly affect the reward function and relate to the early termination conditions.
By only including quantities that are critical to the reward function and termination, the agent is able to focus on the most relevant aspects of the task.
To ensure the agent begins each episode from a feasible state, the initial conditions include a small vertical velocity and displacement.
The agent alone predicts the first action that is executed and used to initialize the transformer-generated sequence.
Because the transformer is not initialized by the first step, the distribution parameters are not present, and their position in the state is padded with sevens.
\begin{table}[H]
\caption{DRL agent state space.}
    \centering
    \begin{tabular}{c l l}
    \toprule
         Quantity & Description & Initial Value or Token \\
         \midrule
         $y$ & Altitude & \unit[per-mode = symbol]{\qty{0.1}\meter} \\
         $V_x$ & Horizontal velocity & \unit[per-mode = symbol]{\qty{0.1}\meter\per\second} \\
         $V_y$ & Vertical velocity & \unit[per-mode = symbol]{\qty{0.1}\meter\per\second} \\
         \midrule
         $\mu_P$ & Mean power & $7$ \\
         $\mu_\theta$ & Mean wing angle & $7$ \\
         $\sigma_P$ & Standard deviation of power & $7$ \\
         $\sigma_\theta$ & Standard deviation of wing angle & $7$ \\
         \bottomrule
    \end{tabular}
    \label{tab:state}
\end{table}


The action space~(Table~\ref{tab:action}) contains the power provided to the motors and the wing angle to the vertical.
The action space of the DRL agent differs from the action space of the DRL environment.
In vanilla DRL, the two action spaces are identical, but as mentioned in Sec.~\ref{sec:TDRL}, the transformer-guided DRL agent's action space corresponds to the z-scores of the actions to be executed from the transformer's action distribution.
The executed actions are analogous to the design variables described in~Table~\ref{tab:opt_setup}.
Time is not directly predicted in DRL, but the environment operates on a time step of $0.1$ seconds with a maximum episode length of $40$ seconds.

\begin{table}[H]
\caption{DRL environment action space, consisting of power ($P$) and wing angle to the vertical ($\theta$).}
    \centering
    \begin{tabular}{l|c c}
    \toprule
        Quantity & $P$ & $\theta$ \\
        Range & $\left[1.8, 3.11\right]\times10^5$\unit{\watt} & $\left[0, \frac{\pi}{2}\right]$ \unit[per-mode = symbol]{\radian} \\
        \bottomrule
    \end{tabular}
    \label{tab:action}
\end{table}

\subsection{Verification}
\begin{figure}[H]
    \centering
    \begin{subfigure}[b]{0.45\linewidth}
        \includegraphics[width=\linewidth]{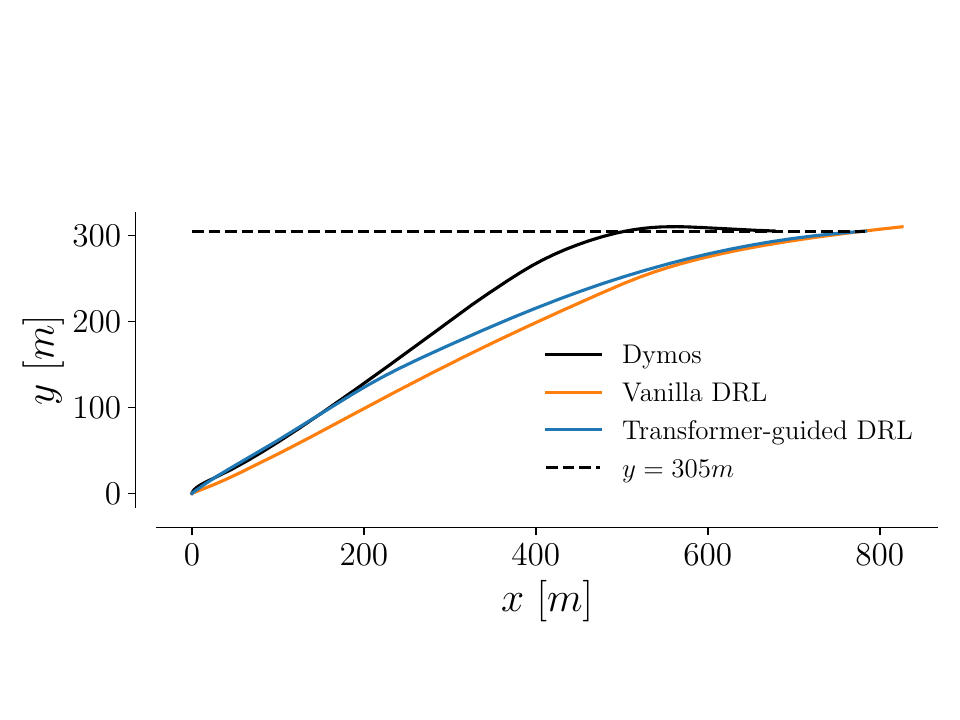}
        \subcaption{}
        \label{fig:trajectories}
    \end{subfigure}
    \begin{subfigure}[b]{0.45\linewidth}
        \includegraphics[width=\linewidth]{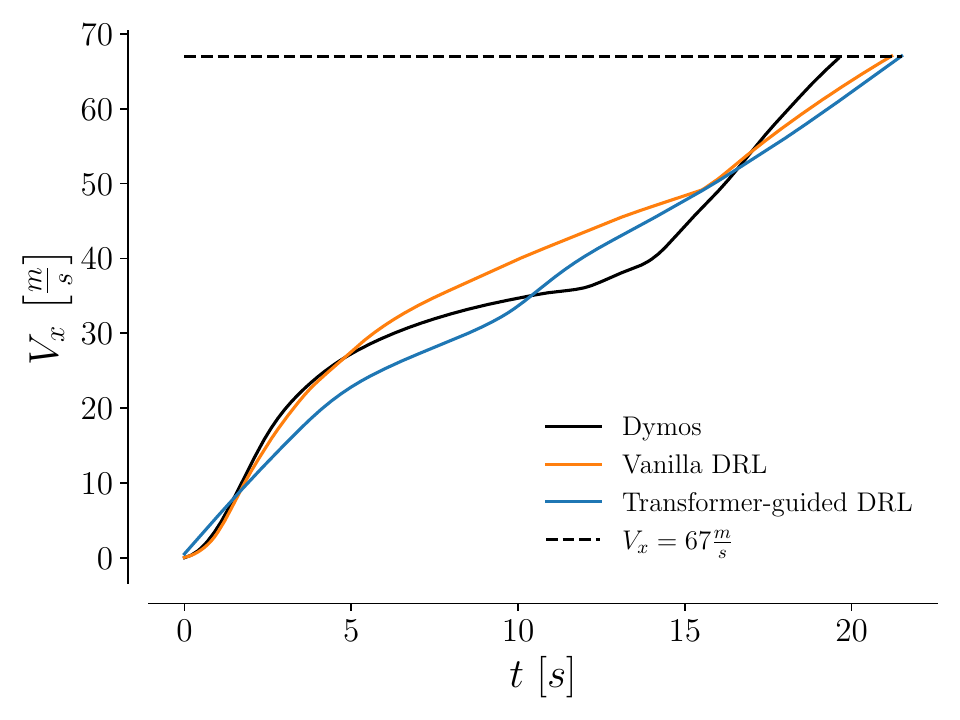}
        \subcaption{}
        \label{fig:velocities}
    \end{subfigure}
    \caption{Comparison of takeoff trajectories: (a) optimal trajectories; (b) horizontal velocity during takeoff.}
\end{figure}

The reference trajectory generated using \texttt{Dymos} consumes 1,693\unit{\watt\hour} over a duration of \unit{\qty{19.6}\second}.
All optimal trajectories (\emph{i.e.}, reference, vanilla DRL, and transformer-guided DRL) satisfy all takeoff conditions, achieving an altitude of \unit{\qty{305}\meter}~(Fig.~\ref{fig:trajectories}) and a horizontal velocity of \unit[per-mode=symbol]{\qty{67}\meter\per\second}~(Fig.~\ref{fig:velocities}).
The trajectory produced by a previous DRL agent trained by~\citet{Roberts2025} consumes 1,759\unit{\watt\hour} over a duration of \unit{\qty{21.4}\second}.
In comparison, the trained transformer-guided DRL agent identifies a trajectory consuming 1,740\unit{\watt\hour} over a duration of \unit{\qty{21.5}\second}.
Moreover, the transformer-guided DRL agent learns to take off with $4.57\times10^6$ time steps, while the vanilla DRL takes $19.79\times10^6$ time steps.
Despite using only one-fourth as many time steps of vanilla DRL, the transformer-guided DRL trajectory exhibits $97.2\%$ accuracy in the minimum energy consumption compared against the \texttt{Dymos} solution~(Table~\ref{tab:energy}).


\begin{table}[h]
\caption{Accuracy on energy consumption for optimized trajectories.}
    \centering
    \begin{tabular}{c|c c}
    \toprule
        & Vanilla DRL & Transformer-guided DRL \\
        \midrule
        \textbf{Accuracy} & $96.1\%$ & $97.2\%$ \\
        \bottomrule
    \end{tabular}
    \label{tab:energy}
\end{table}


\section{Conclusion and Future Work}
\label{sec:conclusion}

Urban air mobility has emerged as a promising solution to alleviate urban congestion, with electric vertical takeoff and landing (eVTOL) aircraft offering quiet and efficient operation in densely populated environments.
A central challenge limiting the widespread deployment of eVTOL aircraft is their high energy demand, particularly during takeoff, which consumes a disproportionate fraction of total energy.
As a result, the design of minimal-energy takeoff trajectories is of substantial interest.
Conventional optimal control methods often struggle with high-dimensional, nonlinear dynamics, while deep reinforcement learning (DRL) can handle such difficulties but typically suffers from training difficulty.

In this work, we developed a transformer-guided DRL framework for optimal takeoff trajectory design of an eVTOL aircraft.
The proposed approach used the sequential-modeling capacity of a transformer---originally developed for machine translation tasks---to guide a DRL agent toward efficient takeoff while alleviating the training complexity that is often encountered in constrained DRL environments.
The transformer, trained on optimal takeoff trajectories for a range of flight conditions, reduced the action space to a more realistic and energy-efficient manifold.

The proposed transformer-guided DRL approach attained an optimal takeoff trajectory with at least $97\%$ accuracy on energy consumption relative to a \texttt{Dymos} reference solution.
Compared to vanilla DRL, the method required only one quarter of the training samples to converge and achieved superior accuracy in minimal energy consumption.
These results demonstrated that incorporating a learned temporal structure into the DRL process improved the training efficiency and the learned policy quality.

Future work will evaluate the framework under a range of flight conditions and incorporate additional path and safety constraints.
Further research will also consider alternative DRL or transformer designs.
The integration of transformer-based sequential modeling with DRL presents a promising pathway toward data-efficient dynamic control strategies for eVTOL trajectories.

\bibliography{sample}

\end{document}